# IDEAL: A software package for analysis of influence diagrams

Sampath Srinivas
srinivas@rpal.com

Jack Breese
breese@rpal.com

Rockwell Science Center, Palo Alto Laboratory
444 High Street
Palo Alto, CA 94301

## Abstract

IDEAL (Influence Diagram Evaluation and Analysis in Lisp) is a software environment for creation and evaluation of belief networks and influence diagrams. IDEAL is primarily a research tool and provides an implementation of many of the latest developments in belief network and influence diagram evaluation in a unified framework. This paper describes IDEAL and some lessons learned during its development.

## 1 Introduction

Over the last few years influence diagrams [5] and belief networks [10] have emerged as attractive representation schema for domains where uncertainty plays an important role. There has been a wealth of work on both basic issues such as the semantics of these representations as well as on efficient algorithms to process them [10,8,14].

This work has now matured to the point where these techniques are finding their way into production systems. IDEAL is a software package that was developed as a platform for research in belief networks and influence diagrams. IDEAL also can be used to create intermediate sized run-time systems and as a library of functions that provides the belief network and influence diagram methodology for embedded use by other applications.

IDEAL incorporates, in a unified framework, many of the latest developments in algorithms for evaluation of belief networks and influence diagrams. In addition, it provides a complete environment for creating, editing and saving belief networks and influence diagrams. In the rest of the paper any reference to 'diagrams' can be taken to refer to influence diagrams and belief networks unless stated otherwise.

## 2 Structure

IDEAL is written in Common Lisp. Lisp was chosen as the implementation language since it is most suited to exploratory programming and quick development. In addition, the software is portable across a wide variety of platforms.

IDEAL is a library of Lisp functions that provides the following features:

- Data structures for representing influence diagrams and belief networks.

- Facilities for creating and editing influence diagrams and belief networks.

- Facilities for copying, saving (to file) and loading influence diagrams and belief networks.

- Utilities that are of use in coding influence diagram manipulation algorithms etc.

- Utilities that provide many useful services like consistency checking and creation of random belief networks.

- Routines that perform some basic transformations of influence diagrams.

- Algorithms for performing inference in influence diagrams and inference and belief propagation in belief networks.

- Influence diagram evaluation algorithms.

These functions can be used interactively by a user typing to a Lisp interpreter or embedded in code by other applications. To preserve portability, IDEAL has only a simple character terminal based user interface. However, it provides hooks for easy development of a graphic interface layered over it on any specific platform. A *graphic interface* has been developed for the Symbolics environment.

# 3 Facilities in IDEAL

## 3.1 Data structures

IDEAL provides abstract data structures for representing influence diagrams and belief networks. These data structures and a tool kit of associated functions provide all the basic low level functionalities required for the creation of belief networks and influence diagrams. This includes creation of directed acyclic graph topologies, creation of probability matrices and other matrices and vectors that are indexed and sized by the states of the nodes in the graph, accessing these matrices and vectors, manipulation of the graph topology, control constructs that allow easy traversal of these node matrices, etc. These are low level features that can be used by programmers to develop functionalities that are not available directly in IDEAL. A user who does not need any additional functionalities can interact with IDEAL with higher level functions described below.

## 3.2 Creating and Editing diagrams

The functions used to create and edit diagrams are at a higher level than the functions that manipulate the low level data structures. These functions expect fully specified diagrams as input and return consistent diagrams after they are done. Some of these functions require interactive input from the user.

Functions to do the following are available: Creation of complete diagrams, adding arcs, deleting arcs, adding nodes, deleting nodes, adding states to a node, deleting states from a node and editing node distributions. These functions make suitable assumptions that guarantee consistency of the diagram after they are done. For example, adding an arc between two nodes extends the distribution of the child node. This extension of the distribution is done such that the child node is independent of the new parent, i.e, the child node has the same distribution given its predecessors regardless of the state of the new node.

Most of these functions can be used embedded in code to create diagrams on the fly. These functions provide the right hooks into IDEAL for a user who is interested primarily in the existing functionality and does not need to go into the low level implementation details.

## 3.3 Copying and Saving Diagrams

The copy function in IDEAL makes a complete copy of a fully specified diagram. This is frequently useful when one wants to make some transformation that might destructively modify the diagram. The copying mechanism provides a means of keeping an unmodified original in the Lisp environment.

IDEAL also has functions that allow the user to save a diagram to file and to reload diagrams from these saved files. IDEAL saves the diagram in text files and so they can easily be exchanged between users at remote sites or on different platforms by electronic mail or other means. The saving function can be made to recognize any extensions that the user may make to the abstract diagram data structures. Thus, any custom information that a user may want to associate with the diagram can also be saved and retrieved.

## 3.4 Utility functions

IDEAL provides a wide variety of utility functions that are of use in conjunction with belief networks and influence diagrams. Consistency checking functions for the following are available: To check whether a diagram is consistent (i.e., it is acyclic, the probability distributions sum to 1, etc), to check whether a diagram is acyclic (a lower level function), to check whether a diagram has a strictly positive distribution and to check whether a diagram is a belief network.

User interface utilities are available for displaying a description of the diagram in text format, for easily accessing nodes in the diagram and for describing the contents of particular nodes of a diagram.

A set of utility functions is available for creating 'random' belief networks. This set of functions is useful for creating examples for testing of belief network algorithms and for quickly creating test belief networks that satisfy certain user defined criteria (for example, see [1]).

In addition to these there are miscellaneous utility functions. Some examples: a function for sorting the nodes in the diagram by graph order and a function that modifies the distributions of a non-strictly positive diagram slightly (as specified by an argument) to make the distribution strictly positive.

## 3.5 Diagram transformations

This is a set of functions, each of which take a consistent diagram as input and return a consistent diagram. These transformations are used in reduction style algorithms [14,13]. They can also be used to make changes in diagrams or to preprocess them before passing them to an inference scheme.

Some of the transformation functions are: Removal of a particular barren node from a diagram, Removal of all barren nodes from a diagram, absorbing a chance node in a diagram, reversing an arc, reducing a deterministic node etc. The transformation functions (as implemented) change the input diagram destructively to yield the result. Details of these transformations can be found in [9,14].

## 3.6 Graphic Interface and documentation

As mentioned before, IDEAL is designed to be a portable tool and so it does not include any implementation specific graphics features. On the other hand, hooks are available for in IDEAL for easily layering a graphics interface over it.

Such an interface has been developed for IDEAL on Symbolics machines. In addition to standard graphic manipulation commands this interface provides most of the functionalities described above either through mouse driven graph manipulation (for eg, reversing an arc) or through convenient menu driven commands. The interface allows convenient access to the Lisp environment in a separate window and can be a very effective programming tool when developing applications based on IDEAL.IDEAL and the Symbolics interface to IDEAL are documented in detail in [17].

## 4 Algorithms in IDEAL

IDEAL provides many different evaluation and inference algorithms. The implementation emphasis is on clarity rather than speed. Each of the algorithms make extensive input checks and also explicitly detects error conditions such as impossible evidence (see Sec 5.2).

The algorithms implemented in IDEAL fall into four classes — reduction algorithms [14,13], message passing algorithms [10,14], clustering algorithms [8,6] and simulation algorithms [10]. The algorithms in each class are closely related to each other but differ in complexity or are applicable to only specific kinds of belief networks. Reduction algorithms are used for influence diagram evaluation (i.e., solving an influence diagram for the optimal decision strategy) and for inference. When used for inference they answer specific queries, i.e, they give the updated belief of a specific target node given a set of evidence nodes. The algorithms in the latter three classes (as implemented) can be used only for inference in belief networks. They give updated beliefs for all the nodes in the network given evidence. The data structures for declaring evidence before an algorithm is called and the data structures where the updated beliefs are found after the algorithm has finished running are the same for all algorithms of the latter three classes. So, if need be, the actual algorithm used can be a decision that is transparent to the end user or any calling function which needs an inference mechanism whose details are irrelevant.

## 4.1 Reduction algorithms

Influence diagram evaluation algorithms as described by Shachter [14] and Rege and Agogino [13] are available. Inference algorithms applicable to both influence diagrams and belief networks are also available as described in the same sources.

These algorithms operate by making a series of transformations (see above) to the input diagram. The input diagram is destructively modified.

## 4.2 Message passing algorithms

Message passing algorithms model each node as a processor that communicate by means of messages. A distributed algorithm from Pearl that applies to polytrees [10] is implemented in IDEAL. This implementation also utilizes work by Peot and Shachter [11]. A conditioning algorithm that works for all belief networks is also available. The conditioning algorithm calculates cutset weights as described by Suermondt and Cooper [18]. A variation of the conditioning algorithm from Peot and Shachter [11] is also available. The conditioning algorithms find cutsets as described by Suermondt and Cooper [19].

## 4.3 Clustering algorithms

Clustering algorithms aggregate the nodes in a belief network into a *join tree* of 'meta' nodes and then run an update scheme on this tree. The updated beliefs for each of the belief network nodes is then calculated from the 'meta' nodes.

IDEAL implements two variations of the basic clustering algorithm described by Lauritzen and Spiegelhalter [8]. The first considers the join tree as a 'meta' belief network and runs a variation of the polytree algorithm [10] on it. The second variation uses an update scheme that operates on clique potentials as described by Jensen et al [6].

Two methods are available for making the fill-in for use in construction of the join tree — Maximum Cardinality Search [20] and a heuristic elimination ordering heuristic from Jensen et. al. [7,12].

## 4.4 Simulation Algorithms

IDEAL implements a simulation algorithm from Pearl [10]. This implementation can only operate on belief networks with strictly positive distributions.

## 4.5 Estimator functions

IDEAL provides run time estimator functions for some of the algorithms implemented in it. Given an algorithm and a particular belief network with a particular state of evidence, the estimator function gives a quick estimate of the complexity of the update process.

In general, belief net inference algorithms consist of two kinds of operations. The first kind are graph operations that are polynomial in the number of nodes in the graph (eg, triangulating a graph for clustering, conversion of a multiply connected network into a singly connected network by instantiating a cutset)[1]. The other class of operations are the actual numerical calculations that are carried out over the probability and potential matrices associated with the graphs. We will refer to this as the update process. The overall exponential complexity algorithm derives from the fact that these matrix operations carried out during the update process take exponential time. The estimator functions in IDEAL give a quick estimate of the complexity of these matrix operations.

The complexity count that is returned is a count of the number of steps the algorithm will spend in spanning the state spaces of the nodes or cliques involved. For example, if a binary node $A$ has a lone binary node $B$ as a predecessor then the complexity count of setting the probability distribution of $A$ is four since one has to cover a state space of $2 \times 2$ states. The complexity of normalizing the belief vector of $A$ is again 4 since one has to cover the state space of the node $A$ twice, once for summing the beliefs and once for normalizing them.

An estimator function for a particular algorithm takes an inference problem as input, i.e, a belief network and associated evidence. The estimator performs the polynomial time graph manipulations that are necessary for initialization before the actual update process can begin. It then applies embedded knowledge of the update process to give an exact count of the number of steps that the update process will take. A step is defined as explained in the previous paragraph. This estimate is made in linear time. So overall, the estimator function runs in time polynomial in the size of the input.

---

[1] Here we refer to the actual graph algorithm implemented as against the algorithm which would give optimal results. For example, the algorithm implemented in IDEAL for finding a loop cutset for conditioning runs in polynomial time while the problem of finding the minimal loop cutset is NP-hard (see [3] for both results).

We have calibrated the estimates yielded by these functions against actual time measurements of how long it takes to solve the corresponding problems. The correlations have been strong (see Sec 5.1).

## 5 Discussion

IDEAL has been a success from the experimental point of view. It has been used both for in-house applications and research both within and outside Rockwell. Some examples of the uses of IDEAL include a decision aiding model for pilots that helps to sort the vast flow of information that comes to the cockpit from the sensors on the plane, a life cycle costs analysis system for Rocket engines, embedded use in a natural language system for story understanding [2] and an implementation of interval influence diagrams [4].

One of the lessons we learned in the process of implementing IDEAL was that many of the algorithm papers do not describe the algorithms in standard algorithmic style. In addition they leave many details incompletely specified. From an engineering point of view, it would be very useful if we had both a more complete description of algorithms and in a more conventional style. IDEAL's emphasis on code readability and explicitness were of great help in detecting and correcting any problems that came up.

### 5.1 Estimator functions

As explained in the previous section, the estimator functions carry out the polynomial time graph manipulations that precede the update process and then give an estimate of the complexity of the update process. The results of the graph manipulation are required to make the estimate. The actual estimate is the result of applying a formula to the results of the graph manipulation. These formulae were derived by analysis of the update process of each algorithm. The estimator functions in IDEAL apply only to exact algorithms (as opposed to approximation algorithms).

As an example of an estimator function, consider the estimator for the Jensen method [6] of clustering. Given a belief network the complexity of initializing the join tree by the Jensen method if given by:

$$\sum_{U \in J} (3 + N(U))S(U)$$

where $U$ represents a Bayesian belief universe, $J$ is the join tree made up of Bayesian belief universes, $N(U)$ is the number of neighbors of $U$ in the join tree and $S(U)$ is the size of the joint state space of the belief network nodes that are members of $U$.

This formula is easily derived as follows. For each belief universe the potential distribution has to be set up by multiplying the distributions of the component belief network nodes. This has complexity $S(U)$.

When a belief universe absorbs from its neighbors the complexity of the operation is $S(U)$. When it updates a neighboring sepset, again the complexity of the operation is $S(U)$. During the collect-evidence operation, each universe absorbs from its 'child' neighbor sepsets and then updates its 'parent' neighbor sepsets. Thus, for each universe the complexity of the operation is $2S(U)$.

During the distribute-evidence operation, each universe first absorbs from its 'parent' sepset neighbor and then updates all the 'child' sepset neighbors. The complexity of the operation is $N(U)S(U)$ for each universe $U$. Summing the terms for initialization of the join tree, the collect-evidence operation and the distribute-evidence operation gives the complexity formula above.

An approximate formula that gives the complexity of the update process in the Jensen algorithm is:

$$\sum_{U \in J}(2 + N(U))S(U) + \sum_{i \in B}[S(U_i) + S(i)]$$

where $U_i$ is the smallest universe (in terms of state space size) that contains node $i$ of the network.

The update process consists of one collect evidence-operation, one distribute-evidence operation and a marginalization operation for setting the belief vectors of the belief network nodes. These factors add up to the formula above. The formula does not take into account the fact that some optimization can be made based on the position of evidence in the join tree. It also does not include the operations needed to declare evidence in the join tree. However, leaving out these terms does not introduce significant error.

We have obtained excellent correlations between the complexity estimates given by the estimator functions for various algorithms and the actual run time. Fig 1 demonstrates the correlation for the update phase of the Jensen algorithm. The data in the graph was collected by running tests on randomly created belief networks.

As expected, particular algorithms suit particular types of problems well. When choosing what algorithm to use, in addition to the type or size of problem, one needs to consider whether the belief network involved needs to be solved just once or solved multiple times with different evidence sets. Conditioning algorithms are competitive (though not necessarily faster) when the problems needs to be solved only once. This could be the case, for example, in a system that constructs belief networks dynamically and uses each network only once. When the same network is used repeatedly with different evidence pieces, the clustering algorithms are superior. The construction of the join tree can be considered as a compilation step of the belief network that needs to be carried out only once.

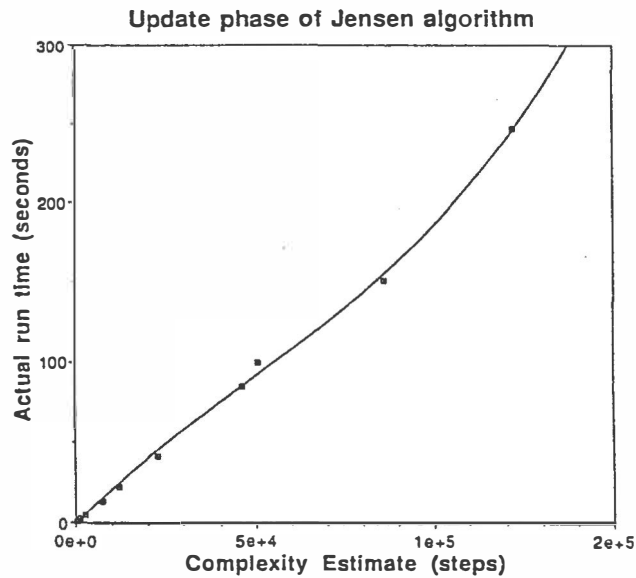

Figure 1: Performance of estimator functions: An example

Though IDEAL is an experimental tool it gives reasonable response times for medium size problems. As an example, a 50 node network developed as part of a decision aid system for aircraft pilots takes about 17 seconds to solve on a Symbolics 3645. IDEAL's speed is limited both by the choice of implementation language and its implementation style, where explicit code rather than speed has been the top priority.

## 5.2 Handling determinacy and inconsistency

In all the algorithms, gains can be made by explicitly detecting determinacy in the network. This can be done as a pre-processing step [15](in which case the network topology itself is modified) or, more generally, in the propagation phase of the algorithm.

When the joint probability distribution of a belief net (i.e, the joint distribution of all the variables in the belief net) is not strictly positive it means that some particular configuration of the belief net is impossible. This in turn implies that some subset(s) of nodes of the belief net have non-strictly positive joint distributions, i.e., the unconditional probability of some joint state of the subset is zero. The actual makeup of these subsets depends on the conditional independencies in the network.

Let the network $I$ (or some subset of nodes of the network) have an impossible state $I = i$. Then, obviously, any conditional probability distribution $P(X/I = i)$ where $X$ is another subset of nodes of the network cannot be assigned meaningfully. If an implementation of any probabilistic inference algorithm does not account for such circumstances, this leads to a divide by zero error if the implementation tries to calculate the distribution $P(X/I = i)$. This occurs either when calculating $P(X/I = i)$ as $P(X, I = i)/P(I = i)$ or when normalizing the representation of $P(X/I = i)$, say $R(X/I = i)$ for all states $x$ of $X$ where each $R(X = x/I = i)$ has been found to be zero. Note that the representation is inconsistent and cannot represent a conditional probability distribution that sums to 1.

An impossible state can occur due to two things:

1. <u>Inconsistent Evidence:</u> The evidence that the user has declared may be inconsistent with the belief net. Let us say that the probabilities encoded in the belief net are such that for a subset of nodes $A$ of the belief net $P(A = a)$ is zero where $a$ is some joint state of the nodes $A$. If the evidence we declare happens to be exactly $a$ or some superset of it (i.e $a$ plus evidence for some nodes outside $A$) then obviously we will hit a divide by zero error when performing inference to find some distribution $P(B/A = a)$ where $B$ is some other set of nodes in the belief net. This is because the distribution we are seeking is hypothetical, unassignable or meaningless, depending on how we look at the problem.

2. <u>Nature of algorithm:</u> An impossible state may also be caused by the nature of the inference algorithm. Consider the conditioning algorithm, for example. It performs whatever inference we are interested in conditioned on every possible joint state of a set of cutset nodes which make the belief net singly connected. The results obtained from each of these conditionings are then 'weighted' to get the results. Thus if the cutset is $A$ and the evidence is $E = e$ and the target node(s) is $B$ then we find $P(B/A = a, E = e)$ for all possible states $a$ of $A$ and then weight these results. If $P(A = a, E = e)$ is zero for some state $a$ of $A$ it is easy to see that we have an impossible state which would lead to a divide by zero error when calculating $P(B/A = a, E = e)$. Thus, in general, an algorithm can hit an impossible situation (which cannot be attributed to inconsistent evidence) if the algorithm calculates any conditional distribution in which the conditioning node set consists of some belief net nodes which are *not* evidence nodes.

### 5.2.1 Reduction algorithms

In reduction algorithms a divide by zero error can occur when we try and find new conditional distributions. This happens only during arc reversal and node absorption. In inference algorithms node absorption is just a special case of arc reversal and so we need to look only at arc reversal.

When performing arc reversal to find a new distribution $P(A/B = b)$ where $A$ is a single node and $B$ is a set of nodes the basic method is to marginalize $P(A, B = b)$ and then normalize it using the marginal. We hit a divide by zero error if the marginal $P(B = b)$ happens to be zero. In such a case IDEAL makes $P(A/B = b)$ a uniform distribution. This is justified because any subsequent manipulation of the distribution $P(A/B = b)$ by a reduction algorithm always involves multiplying it into $P(B = b)$ first. We know that $P(B = b)$ is zero and so $P(A/B = b)$ can be anything. The advantage of this uniform assignment is that the diagram remains consistent (i.e., the numbers still constitute a valid probability distribution) even after the transformation. The disadvantage is that if the user's query to the system was $P(A/B = b)$ and $P(B = b)$ happens to be zero for some state of $B$ then the user will not realize it and may ascribe some meaning to the distribution $P(A/B = b)$ even though it has no meaning. Note that this effectively amounts to outputting garbage when the evidence is impossible (the evidence being that particular state $b$ of $B$).

### 5.2.2 Message passing algorithms

The polytree algorithm, as implemented in IDEAL, cannot hit the divide by zero error during the propagation phase since it calculates only joint probabilities. However, when normalizing the beliefs of each belief network node after the propagation is done, it is possible to find that the marginal is zero. This directly implies that the evidence declared before the propagation is impossible (i.e., $P(E = e) = 0$) since the marginal is nothing but $P(E = e)$. IDEAL detects this situation explicitly and tells the user that the evidence is impossible.

This conditioning algorithm makes the belief net a poly tree by clamping the states of a cycle cutset of nodes $S$. The evidence is propagated as by the polytree algorithm for each of the evidence pieces and then the result is weighted to get the beliefs of each node given the evidence alone.

IDEAL supports two conditioning implementations. The first calculates cutset weights explicitly. In other words, for every node $A$ we calculate $P(A/S = s, E = e)$ and then use that to calculate $P(A/E = e)$ as the marginal of the product $P(A/S = s, E = e)P(S = s/E = e)$, where $P(S = s/E = e)$ is a 'mixing' probability. We will hit the divide by zero error when $P(S = s, E = e)$ is zero and we try and calculate $P(A/S = s, E = e)$.

In this implementation, a cutset conditioning case $s$ for which $P(S = s, E = e) = 0$ does not contribute to the overall belief. So to avoid an error the cutset algorithm checks for the occurrence of $P(S = s, E = e) = 0$ during the recursive update process that determines $P(S = s/E = e)$. If the condition occurs then that cutset conditioning case $s$ is skipped. Other than being a graceful technique to detect an impossible situation, this step, in conjunction with Suermondt and Cooper's [19] technique for calculating cutset weights, can lead to substantial complexity gains since whole classes of impossible cutset cases can be detected and skipped with very little effort. For example, if the cutset consists of three binary nodes $A$, $B$ and $C$ (in graph order $(A, B, C)$), then knowing that $P(A = t) = 0$ immediately eliminates 4 cutset cases, one for each state combination of $B$ and $C$ in conjunction with $A = t$.

In the second conditioning implementation [11] no conditional probabilities are calculated during the propagation phase and so no divide by zero errors are possible. However, it is possible that when marginalizing the belief vectors of the nodes after the propagation, the marginals are zero. This implies that the evidence that has been propagated is impossible (see previous subsection). IDEAL detects this situation explicitly in both conditioning implementations.

### 5.2.3 Clustering Algorithms

IDEAL supports two clustering algorithm implementations. The first implementation creates a join tree of cliques and calculates the conditional probabilities in the join tree. Consider a clique $A$ with a parent clique $B$. We hit the divide by zero error when $P(B = b)$ is 0 and we try and calculate $P(A/B = b)$. When creating the join tree we assign $P(A = a/B = b) = 0$ (we could assign anything, in fact) for all states $a$ of $A$ when $P(B = b) = 0$. After the join tree is created the clustering algorithm uses a variant of the polytree algorithm for evidence propagation and so the divide by zero problem cannot come up.

The second implementation from [6] handles a divide by zero condition during the propagation as described in the original paper. After propagation, if a zero marginal is encountered when normalizing the beliefs this implies that the evidence was impossible. IDEAL signals the fact explicitly in both clustering implementations.

### 5.2.4 Simulation Algorithms

The simulation algorithm coded in IDEAL *cannot* handle non-strictly positive belief networks. If such a belief network is given as input the algorithm breaks with an appropriate warning.

## 6 Further developments

We foresee more work on developing efficient estimator functions. Each estimator function may be expanded into a class of functions where one may trade off the accuracy of the estimate with the time required to make the estimate. It may be possible to use these estimator functions to help choose between competing algorithms for a given problem or to use them as a search function to search through a space of competing alternative solutions.

IDEAL, has incorporated almost all the published work to date on exact belief network and influence diagram algorithms. We will probably include any promising new methods that come up (for example, nested dissection [3]) so that we can choose the best possible method for the applications we have in mind.

We will also be including some approximation algorithms such as Likelihood weighting [16].

## 7 Acknowledgements

We would like to thank Robert Goldman for being an invaluable source of suggestions, bug reports and enhancements. We would also like to thank Bruce D'ambrosio, Keiji Kanazawa, Mark Peot and other users of IDEAL for their suggestions and help.